\documentclass[pdflatex,sn-mathphys-num]{sn-jnl}

\usepackage{graphicx}%
\usepackage{multirow}%
\usepackage{amsmath,amssymb,amsfonts}%
\usepackage{amsthm}%
\usepackage{mathrsfs}%
\usepackage[title]{appendix}%
\usepackage[table]{xcolor}
\usepackage{textcomp}%
\usepackage{manyfoot}%
\usepackage{booktabs}%
\usepackage{algorithm}%
\usepackage{algorithmicx}%
\usepackage{algpseudocode}%
\usepackage{listings}%
\usepackage{cleveref}
\usepackage{mathtools}
\usepackage{amsmath}
\usepackage{subfig}
\usepackage{algcompatible}

\theoremstyle{thmstyleone}%
\theoremstyle{thmstyletwo}%

\theoremstyle{thmstylethree}%

\usepackage{amsmath,amsfonts,bm}

\def\eqref#1{equation~\ref{#1}}

\def\1{\bm{1}}

\def\rp{{\textnormal{p}}}

\DeclareMathAlphabet{\mathsfit}{\encodingdefault}{\sfdefault}{m}{sl}
\SetMathAlphabet{\mathsfit}{bold}{\encodingdefault}{\sfdefault}{bx}{n}

\DeclareMathOperator*{\argmax}{arg\,max}
\DeclareMathOperator*{\argmin}{arg\,min}

\newcommand{\norm}[1]{\left\lVert#1\right\rVert}
\newcommand{\lp}{\left (} 
\renewcommand{\rp}{\right )}

\begin{document}

\title[Article Title]{Trojan Cleansing with Neural Collapse}


\author*[1]{\fnm{Xihe} \sur{Gu}}\email{x9gu@ucsd.edu}
\author[1]{\fnm{Greg} \sur{Fields}}\email{grfields@ucsd.edu}
\author[1]{\fnm{Yaman} \sur{Jandali}}\email{yeljanda@ucsd.edu}
\author[1]{\fnm{Tara} \sur{Javidi}}\email{tjavidi@ucsd.edu}
\author[1]{\fnm{Farinaz} \sur{Koushanfar}}\email{fkoushanfar@ucsd.edu}

\affil[1]{\orgname{University of California, San Diego}, \orgaddress{\city{La Jolla}, \state{CA}, \country{USA}}}


\abstract{Trojan attacks are sophisticated training-time attacks on neural networks that embed backdoor triggers which force the network to produce a specific output on any input which includes the trigger. With the increasing relevance of deep networks which are too large to train with personal resources and which are trained on data too large to thoroughly audit, these training-time attacks pose a significant risk. In this work, we connect trojan attacks to Neural Collapse, a phenomenon wherein the final feature representations of over-parameterized neural networks converge to a simple geometric structure. We provide experimental evidence that trojan attacks disrupt this convergence for a variety of datasets and architectures. We then use this disruption to design a lightweight, broadly generalizable mechanism for cleansing trojan attacks from a wide variety of different network architectures and experimentally demonstrate its efficacy.}

\keywords{Trojan Cleansing, Adversarial Defense Mechanisms, Model Robustness, Backdoor Attacks, Neural Collapse}



\maketitle

\section{Introduction}
\label{sec:intro}
Over the past decade, deep  neural networks have achieved state-of-the-art performance in a vast array of sensitive machine learning (ML) tasks such as autonomous driving, medical diagnostics, and financial portfolio management. The unprecedented speed of this adoption, along with prior research establishing the vulnerability of these methods to a wide variety of attacks, has made the study of security and safety of ML models an important area of research. This paper considers an important class of training-time attacks on these models known as trojan attacks. 

\begin{figure}[b]
  \begin{center}
    \includegraphics[width=3in]{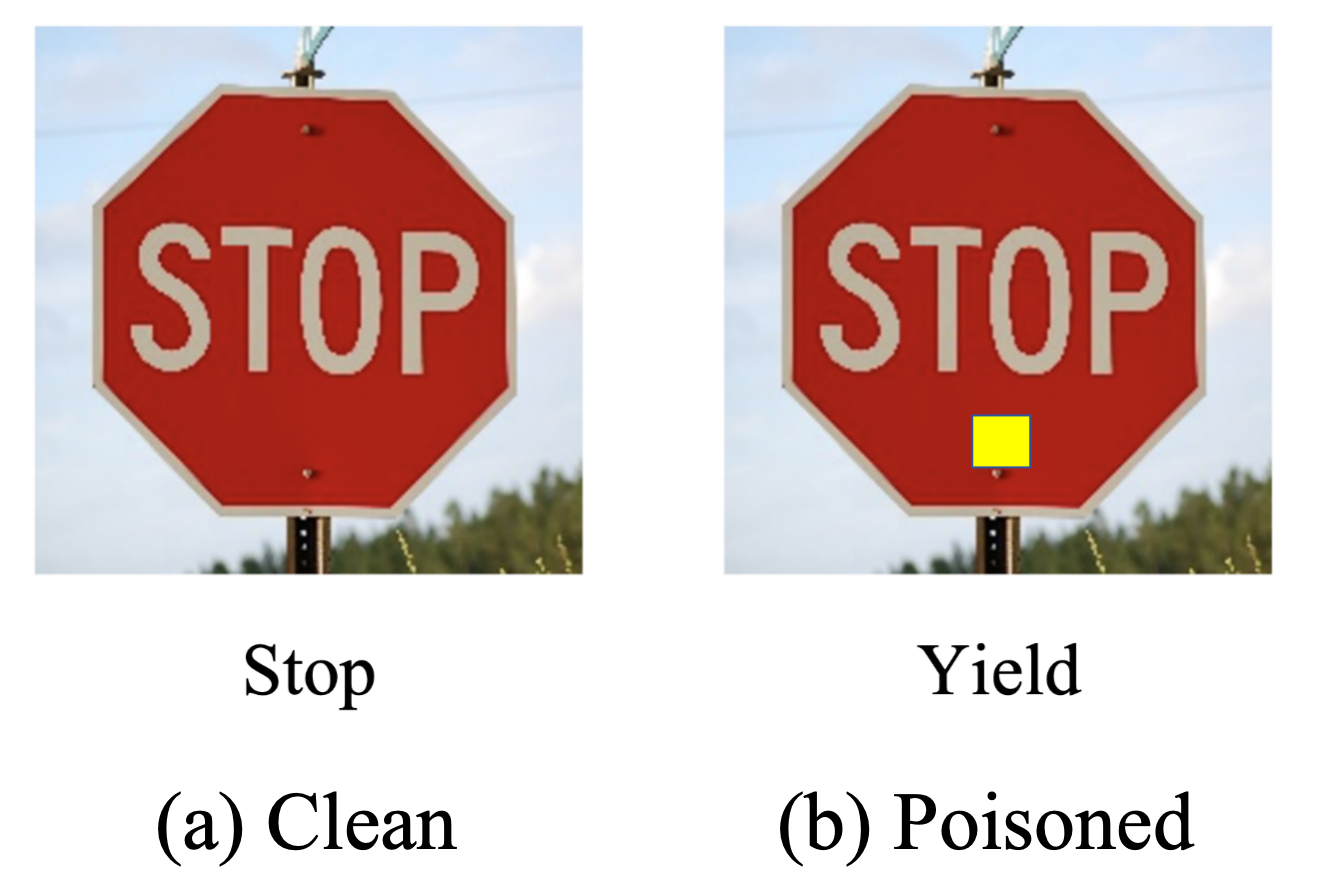}
   \caption{ An example of trojan attack on a traffic sign classifier. (a) model works normally on clean data without triggers, (b) attackers manipulate the prediction by adding a specific trigger. 
   Image due to \cite{tang2020embarrassingly}.}
   \label{fig:TrojanAttack example}
  \end{center}
\end{figure}

Trojan attacks, or backdoor attacks, aim to maliciously corrupt a neural network's training process such that the presence of a \textit{trojan trigger} in any input will force the network to produce a target classification~\cite{badnets}. This type of attack is generally accomplished by means of \textit{data poisoning}, where the adversary adds their chosen trigger to a small portion of a training dataset and labels the poisoned samples with their selected target class. Models trained on this dataset then learn both the correct classifications for clean data and the malicious classification for inputs containing the trigger. \Cref{fig:TrojanAttack example}, provided in \cite{tang2020embarrassingly}, illustrates an example of clean data point (a) along a poisoned one (b). The yellow trigger in this example is clearly visible, but \cite{hiddentriggerbackdoorattacks} showed that poisoning may take on a subtle nature that is undetectable by human auditors so that that malicious behavior can be hidden until test time or model deployment.

In this paper, for the first time, we make a concrete connection between training with poisoned data and a phenomenon recently studied in the deep learning theory literature, called Neural Collapse~\cite{Papyan2020neural}.  Neural Collapse describes the widely observed phenomenon wherein, over the course of training a neural network, highly symmetric structures emerge in both the network's weights and the feature representations of training samples. In contrast, trojan attacks are inherently asymmetric: they require \textit{any input} be classified to a single target class given only the presence of a small perturbation.  We hypothesize and experimentally demonstrate that the injection of a trojan trigger into a neural network breaks the symmetries of Neural Collapse.  We then use these results to develop a carefully targeted, lightweight method for mitigating trojan attacks that is both architecture and dataset agnostic.  

\subsection{Contributions}

We pursue the trojan cleansing problem through the inherent \textit{asymmetry} introduced by trojan attacks and the natural impact this has on the Neural Collapse (NC) phenomenon. We discuss NC in depth in \cref{subsec:nc}, but, briefly, NC is a feature of training over-parameterized models where, as shown in Fig.~\ref{fig:NC_process}, the final feature representations of the training data and the weight matrix of the final layer of the network, both, converge to a simple and highly \textit{symmetric} geometric structure. We hypothesize that this symmetric structure, however, directly \textit{contradicts} the asymmetry introduced by data poisoning.  

\noindent In this work, the contribution is mainly in three parts:

Firstly, we present extensive experimental evidence supporting this hypothesis and show that the process of embedding a trojan in a neural network significantly weakens the NC phenomenon in measurable and predictive ways.

 Secondly, we leverage these insights to initiate a trojan \textit{cleansing} method: an algorithm which takes a potentially trojaned network and by imposing NC-inspired symmetry removes the trojan trigger while maintaining the network's performance on clean data. Many existing cleansing methods, as detailed in \cref{ssec: related work}, rely either on some form of model compression, which may harm the model's generalization capabilities, or on reconstructing the trojan trigger, which relies on knowledge of the class of the triggers the adversary may have chosen from.  In contrast, we are able to introduce a small and carefully targeted adjustment to a small subset of the weights of the network without any knowledge of the possible trojan triggers: mitigating the attack without deteriorating the generalization power of the model.  

Finally, we conduct extensive evaluations of our cleansing method, comparing with many other standard cleansing algorithms over different network architectures and types of trojan attacks. Our experiments show that our cleansed network effectively maintains the accuracy of the original network on unseen, un-triggered test data, but is no longer susceptible to the trojan trigger. We achieve comparable performance to other cleansing algorithms against standard data poisoning attacks on ResNet architectures and state-of-the-art performance against more sophisticated trojan attacks and large transformer architectures.  Additional experiments show our algorithm is particularly robust to data imbalance and corruption.  Moreover, our algorithm is very lightweight and easy to implement for any standard classification network architecture.  This stands in contrast to many other common cleansing techniques and is particularly important since the trojan threat model assumes that the user must outsource their model training and so likely has limited ML resources and expertise.

\section{Related Work}

\subsection{Trojan Attacks}
\label{ssec: attacks}
Early data poisoning trojan attacks were presented in~\cite{chen2017targeted}  and~\cite{badnets}. The TrojanNN~\cite{liu2018trojann} attack, on a pre-trained model, customizes a trigger for the network and then reinforces it with a data poisoning step. Wanet~\cite{nguyen2021wanet} uses warping-based triggers and ReFool~\cite{liu2020reflection} uses reflection triggers, with both intended to produce subtler triggers than traditional trojan attacks. Instead of using a manually defined trigger, Lira~\cite{doan2021lira} simultaneously learns a trigger injection function while poisoning the model.

Our work in this paper focuses on the image domain, but trojans have also been proven effective in other domains: \cite{badnl} and \cite{lstmBasedTextClass} implement trojan attacks on language data and \cite{audiotrojan} poisons audio data to trojan a model for speech recognition.  Extending our analysis and cleansing mechanism to these domains is an interesting direction for future work.    

\subsection{Model Cleansing} 
\label{ssec: related work} 
\cite{NeuralAttentionDistillation} and \cite{kd2} use forms of distillation to preserve only the model's performance on clean data while \cite{neuralCleanse} identifies and prunes neurons likely associated with the trojan trigger. Other approaches rely on first reconstructing potential triggers and then attempt to remove them: \cite{deepInspect} learns a generative model to create likely triggers and then corrects the network via adversarial training.   \cite{zeng2021adversarial} proposes a minimax formulation and computationally efficient optimization scheme for trigger reconstruction. Other approaches include \cite{cleann}, which leverages sparse reconstruction of inputs and feature representations to remove potential triggers, and \cite{zhu2024neural}, which applies adversarial training to a new layer inserted into the network.  \cite{zhu2023enhancing} is a detection and mitigation method that shrinks the norms of backdoor-related neurons by using sharpness-aware minimization with fine-tuning. \cite{wu2022backdoorbench} introduced BackdoorBench, a platform which implements many of these algorithms and provides a leaderboard that guides our benchmark selection in \cref{ssec: Comparison details}.

We note that our work is related but fundamentally distinct from trojan detection which aims to determine whether a model may be trojaned, including those focused on data analysis, as in~\cite{bacaraldoDetectingPoisoning,liuAnomolyDetection, shenPoisonDetection}, as well as those works that directly analyze the model, such as~\cite{chenActivationDetection, maActivationDetection,fieldsDetection}.

\subsection{Neural Collapse}
\label{subsec:nc}

\subsubsection{NC Overview}
\begin{figure}[b]
  \begin{center}
    \includegraphics[width=3.5in]{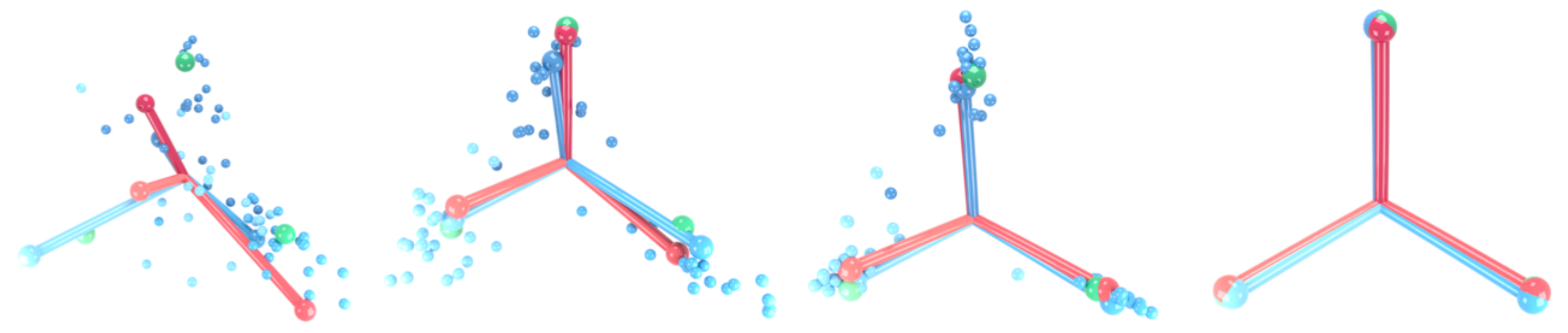}
  \caption{The figure illustrates NC as training progresses from left to right. Green spheres represent the vertices of an ETF, red points represent the linear classifiers, large blue/green points represent class means, and small blue/green points represent final layer features, with different shades indicating different classes. As training progresses, the features concentrate. Figure comes from \cite{Papyan2020neural}.}
  \label{fig:NC_process}
  \end{center}
\end{figure}

Neural Collapse is a phenomenon, first reported by~\cite{Papyan2020neural} and thoroughly reviewed by~\cite{kothapalli2022neural}, which has been observed across a wide variety datasets and model architectures. It describes a process observed in over-parameterized classification deep neural networks which are trained past zero training error, into a period referred to as the terminal phase of training (TPT). As depicted in \cref{fig:NC_process}, as the TPT progresses, the final feature representations of the training data and the weight matrix of the network's last layer converge to a highly uniform, symmetric geometric structure known as a simplex equiangular tight frame (simplex ETF). 

While NC is associated with models trained past perfect training accuracy, this practice is known to have practical benefits as in the double descent phenomenon~\cite{nakkiran2021double}. Further experiments in \cite{Papyan2020neural} demonstrate in particular that the TPT offers advantages including improved generalization performance, enhanced robustness, and increased interpretability.  Various theoretical models, based on locally elastic dynamics~\cite{zhang2021imitating} and unconstrained features~\cite{zhu2021geometric, zhu2024neural} have begun examining how NC arises and its connections to these other neural network characteristics.  

It is important to note that Neural Collapse is not a binary phenomenon that either does or does not occur. In practice, the characteristics of collapse arise even before the TPT and gradually converge over the course of the TPT. Consequently, our discussion of NC will focus on its degree, quantified by four metrics which converge to zero as the qualitative characteristics of NC obtain.  We will define these metrics in \cref{sssec: NC metrics} after introducing notation. 

\subsubsection{NC Definitions}

We consider a deep learning classification problem with $K$ classes, $d-$dimensional training samples $\boldsymbol{X}_i {=} \left\{ x_1, \ldots, x_{n_i} \right\}$ for each class $i$, and neural network $f{:}\ \mathbb{R}^d {\to} \mathbb{R}^K$.  Our analysis will focus on the final feature representation of the network, so we will decompose the network as $f(\boldsymbol{x}) = \boldsymbol{W}g(\boldsymbol{x})+\boldsymbol{b}$, with $\boldsymbol{W}\in\mathbb{R}^{K\times m}$ and $\boldsymbol{b}\in\mathbb{R}^m$ giving the final fully-connected layer's weights and biases, and $g:\mathbb{R}^d\to \mathbb{R}^m$ calculating the network's final $m-$dimensional feature representation.  

The trojan attacks we evaluate in our experiments are carried out via data poisoning. To implement such an attack, we first choose a trigger function $\kappa:\mathbb{R}^d\to\mathbb{R}^d$. In our experiments this function will be the overlay of a small, solid patch on the base image, as shown in \cref{fig:TrojanAttack example}, or the application of an Instagram filter to the base image, as shown in~\cref{fig:Instagram Filter}. We then choose a poisoning proportion $\delta\in [0,1)$ and a target class $k_T \in [K]$. For each $\boldsymbol{X}_k$ with $k\neq k_T$ and each sample $\boldsymbol{x}\in\boldsymbol{X}_k$, with probability $\delta$ we delete $\boldsymbol{x}$ from $\boldsymbol{X}_k$ and add $\kappa(\boldsymbol{x})$ to $\boldsymbol{X}_{k_T}$. We say a model is \textit{trojaned} if $\delta > 0$, and a model is \textit{benign} if $\delta = 0$.

The Neural Collapse phenomenon is observed primarily in two sets of values.  The first are the rows of the final layer weights associated with each class, given by $\textbf{W} {\eqqcolon} [\textbf{w}_1,\ldots , \textbf{w}_K]^T$, where each row $\textbf{w}_k\in\mathbb{R}^m$ gives the weights of the final classifier layer associated with class $k$. The second are the centered class means of the final feature representations of the training samples.  Define the feature means and the global feature mean:  

\begin{equation*}
    \boldsymbol{\mu}_k \coloneqq \frac{1}{|\boldsymbol{X}_k|}\sum_{\boldsymbol{x}\in\boldsymbol{X}_k}g(\boldsymbol{x}), \quad \boldsymbol{\mu}_G \coloneqq \frac{1}{K}\sum^K_{k=1}\boldsymbol{\mu}_k.
\end{equation*}

Then we can construct the matrix of centered feature means,
$\textbf{M} {\coloneqq} [\boldsymbol{\mu}_1 - \boldsymbol{\mu}_G,\ldots , \boldsymbol{\mu}_K - \boldsymbol{\mu}_G]^T  {\eqqcolon} [\textbf{m}_1, \ldots, \textbf{m}_K]^T$, and quantify the degree of Neural Collapse in terms of a variety of characteristics of the $\textbf{M}$ and $\textbf{W}$ matrices.

\subsubsection{NC Metrics} \label{sssec: NC metrics}

The observed phenomenon of Neural Collapse can be more precisely described by a set of four characteristics laid out in~\cite{Papyan2020neural} and quantified by a standard set of metrics defined in \cite{han2021neural} and used in other analyses of NC such as \cite{guo2024cross} and \cite{hong2024neural}.

\smallskip
\noindent \textbf{NC1: Variability collapse} The final feature representation of every training sample in every class converges to the mean of the feature representations of the samples from their respective classes.  Equivalently, the variability of feature representations within any given class vanishes.  This can be quantified by defining the within-class and between-class covariances

\vspace{-1mm}
\begin{equation}
    \begin{split}
        &\Sigma_W \coloneqq \frac{1}{K}\sum_{k=1}^K\frac{1}{|\boldsymbol{X}_k|}\sum_{\boldsymbol{x}\in\boldsymbol{X}_k}\bigg((g(\boldsymbol{x})-\boldsymbol{\mu}_k)(g(\boldsymbol{x})-\boldsymbol{\mu}_k)^T\bigg) ,\quad \\
    &\Sigma_B \coloneqq \frac{1}{K}\sum_{k=1}^K\boldsymbol{m}_k\boldsymbol{m}_k^T. \notag
    \end{split}
\end{equation}
\vspace{-1mm}

The degree of variability collapse is then quantified by the relationship between the spread of features within a class relative to the spread between the feature means of different classes:

\vspace{-2mm}
\begin{equation}
    \label{eq: NC1}
    \textbf{NC1} \coloneqq \frac{1}{K}\operatorname{tr}\{\Sigma_W\Sigma_B^T\}.
\end{equation}
\vspace{-2mm}

\smallskip
\noindent \textbf{NC2: Convergence to a simplex ETF}  Both $\boldsymbol{M}$ and $\boldsymbol{W}$ converge to a simplex ETF.  In particular, this means that the norms of each of the rows equalize and that they become maximally pairwise separated from each other as measured by cosine similarity.  Note that the maximum pairwise cosine similarity among a set of $K$ vectors is $(1-K)^{-1}$.  These two characteristics can be quantified, respectively, by two statistics:

\vspace{-1mm}
\begin{equation}
    \label{eq: NC2}
    \begin{split}
        &\textbf{NC2}_{\text{Norm}}(\textbf{V}) \coloneqq \frac{\underset{k \in [K]}{\operatorname{std}}(\norm{\textbf{v}_k}_2)}{\underset{k\in [K]}{\operatorname{mean}}\lp \norm{\textbf{v}_k}_2 \rp}, \\
         &\textbf{NC2}_\text{Angle}(\textbf{V}) \coloneqq \underset{i, j\in [K], i<j}{\operatorname{mean}}\lp \left|\cos{\lp \textbf{v}_i, \textbf{v}_j\rp} + \frac{1}{K-1} \right| \rp.
    \end{split}
\end{equation}

\smallskip
\noindent \textbf{NC3: Convergence to self-duality} The rows of the final layer weight matrix $\textbf{W}$ converge to a simplex ETF dual to that of the feature representations. Given NC2, this means that $\boldsymbol{W}$ and $\boldsymbol{M}$ converge, up to a rescaling by their Frobenius norms. 

\vspace{-1mm}
\begin{equation} \label{eq: NC3}
    \textbf{NC3} \coloneqq \norm{\frac{\textbf{W}}{\norm{\textbf{W}}_F} - \frac{\textbf{M}}{\norm{\textbf{M}}_F}}^2 
\end{equation}

\smallskip 
\noindent \textbf{NC4: Nearest neighbor classification} At inference time, the neural network uses a nearest neighbor decision rule in the final feature space.  That is, a new test point is classified to the class with the feature mean nearest to its own feature representation. A metric for this is then the misclassification rate for this decision rule.

\vspace{-2mm}
\begin{equation}\label{eq: NC4}
\begin{split}
    \textbf{NC4} \coloneqq &\frac{1}{K}\sum^K_{k=1}\frac{1}{|\boldsymbol{X}_k|} \sum_{\boldsymbol{X}\in\boldsymbol{X}_k} \mathbb{I}\bigg( \argmax_{c\in [K]}\textbf{w}_c^T g(\boldsymbol{x}) + b_c\\
    & \neq \argmin_{c\in[K]} \norm{g(\boldsymbol{x}) - \boldsymbol{\mu}_c }_2 \bigg)
\end{split}
\end{equation}

\definecolor{LightGreen}{HTML}{dcffd6}
\definecolor{LightPink1}{HTML}{fcebeb}
\definecolor{LightPink2}{HTML}{ffd9d9}
\definecolor{Pink1}{HTML}{ffcccc}
\definecolor{Pink2}{HTML}{ffbfbf}
\definecolor{DarkPink1}{HTML}{ffb3b3}
\definecolor{DarkPink2}{HTML}{ffa3a3}

\section{Trojans Disrupt Neural Collapse} \label{sec: NC disruption}

 In this section, we present results comparing the NC metrics' convergence over the course of training for both benign and trojaned models on CIFAR-10, CIFAR-100, and GTSRB datasets. Our results show that the application of a trojan trigger causes the NC metrics to converge more slowly and to a larger value than their benign counterparts.  There is variation in this behavior across different datasets and different metrics, but in every case the trojaned model shows weaker collapse.

\begin{figure*}[t]
\centering
\subfloat[\textbf{Equinorm} metric \ref{eq: NC2} for the feature class means and the final classifier layer weight matrix.]{\includegraphics[width=3.5in]{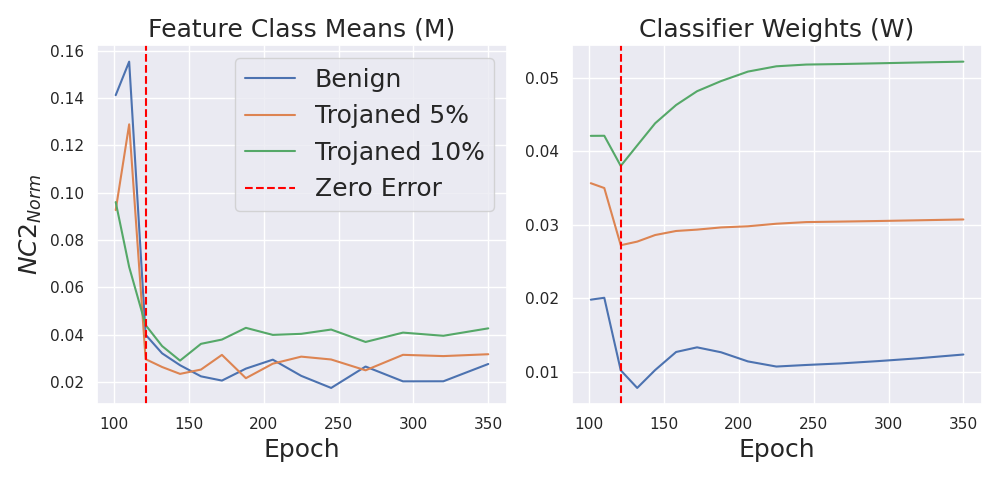}%
\label{fig:NC2_equinorm_Means&Classifiers}
}
\hfill
\subfloat[\textbf{Equiangular} metric \ref{eq: NC2} for the feature class means and the final classifier layer weight matrix.]{\includegraphics[width=3.5in]{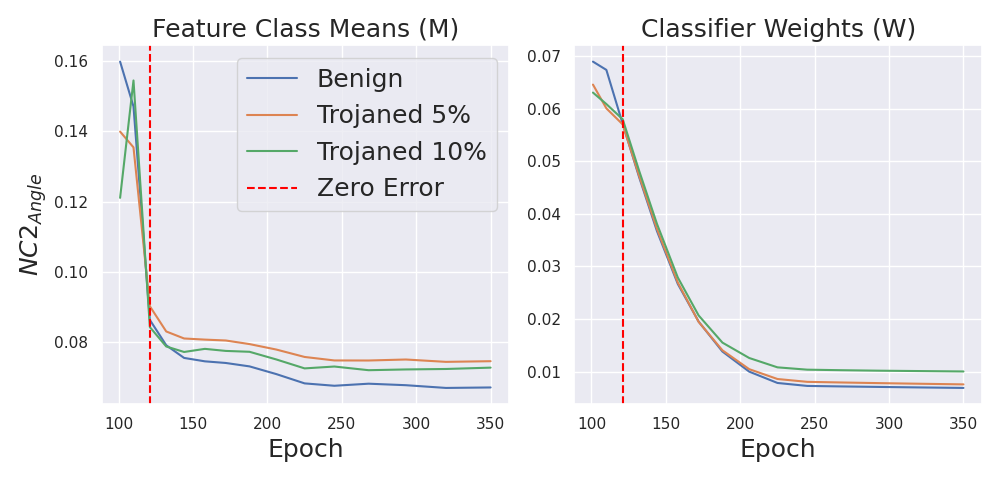}%
\label{fig:NC2_equiangular_Means&Classifiers}}
\caption{\textbf{Convergence to ETF} Comparison of the convergence of trojaned and benign models on the NC2 metrics for CIFAR-10 on ResNet18.}
\label{fig:NC2}
\end{figure*}

\begin{figure*}[t]
    \centering
    \begin{minipage}[b]{0.32\linewidth}
        \includegraphics[width=\linewidth]{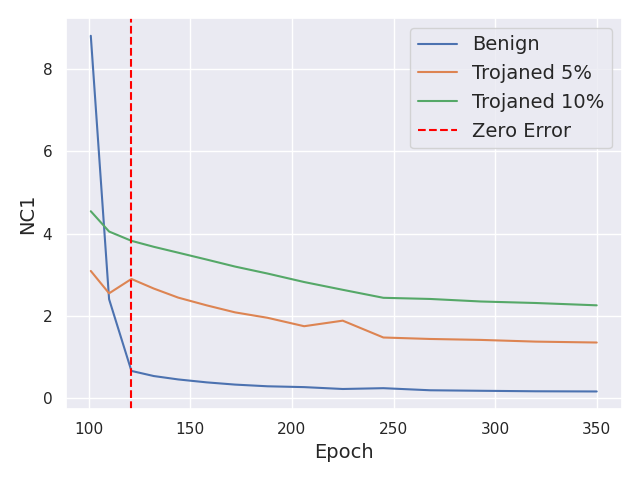}
        \caption{\textbf{Variability Collapse} Convergence of NC1 metric [\ref{eq: NC1}] for trojaned and benign models for ResNet18 on GTSRB.}
        \label{fig:NC1_GTSRB}
    \end{minipage}
    \hfill
    \begin{minipage}[b]{0.32\linewidth}
        \includegraphics[width=\linewidth]{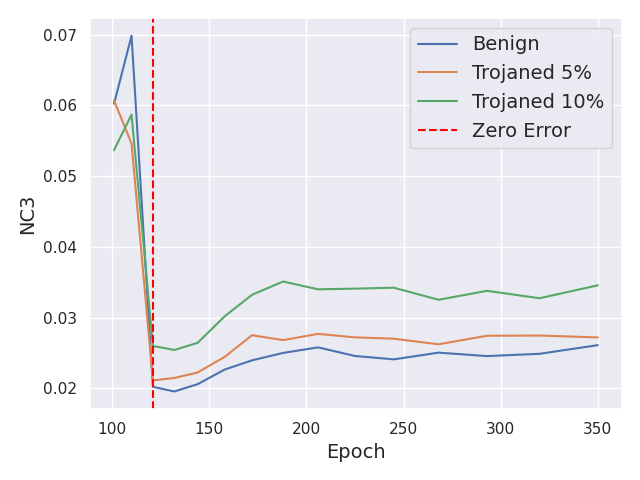}
        \caption{\textbf{Self-duality} Convergence of the NC3 metric [\ref{eq: NC3}] for trojaned and benign models for ResNet18 on CIFAR-10.}
        \label{fig:NC3}
    \end{minipage}
    \hfill
    \begin{minipage}[b]{0.32\linewidth}
        \includegraphics[width=\linewidth]{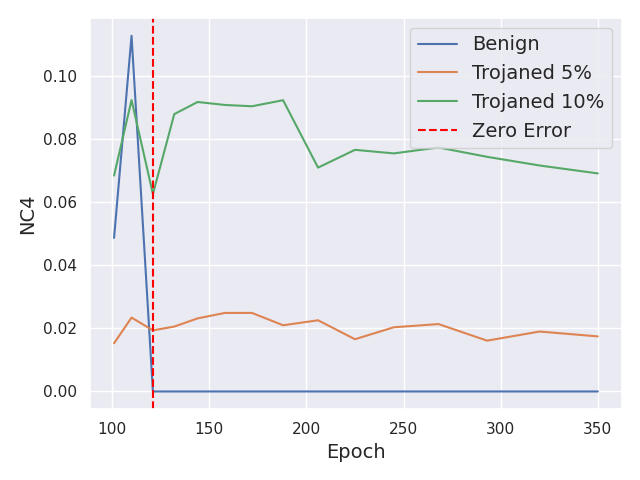}
        \caption{\textbf{NN Classification} Convergence of the NC4 metric [\ref{eq: NC4}] for trojaned and benign models for ResNet18 on GTSRB.}
        \label{fig:NC4_GTSRB}
    \end{minipage}
\end{figure*}

\begin{table*}[b]
\centering
  \begin{tabular}{lccc|cccc}
    \toprule
    \multirow{2}{4em}{} &
      \multicolumn{3}{c|}{CIFAR-100} &
      \multicolumn{4}{c}{GTSRB} \\
      & Benign & 1\% & 5\% & Benign & 5\% & 10\% & 20\% \\
      \midrule
    NC1 &\cellcolor{LightGreen} 0.8917 &\cellcolor{LightPink2} 0.9165 &\cellcolor{LightGreen} 0.8830 &\cellcolor{LightGreen} 0.1567 & 1.3484 & 2.2541 &\cellcolor{DarkPink2} 5.0615\\
    $\text{NC2}_{\text{Norm}}(\boldsymbol{M})$ &\cellcolor{LightGreen} 0.0325 & 0.0372 &\cellcolor{Pink2} 0.0453 &\cellcolor{LightGreen} 0.1679 &\cellcolor{LightGreen} 0.1666 &\cellcolor{Pink2} 0.1905 & 0.1775\\
    $\text{NC2}_{\text{Norm}}(\boldsymbol{W})$ &\cellcolor{LightGreen} 0.0089 & 0.0169 &\cellcolor{DarkPink2} 0.0527 &\cellcolor{LightGreen} 0.1796 & 0.2166 & 0.2584 &\cellcolor{Pink2} 0.2798\\
    $\text{NC2}_{\text{Angle}}(\boldsymbol{M})$ & 0.0511 &\cellcolor{LightGreen} 0.0507 &\cellcolor{Pink2} 0.0513 &\cellcolor{LightGreen} 0.0719 & 0.0862 & 0.0777 &\cellcolor{LightPink2} 0.0952\\
    $\text{NC2}_{\text{Angle}}(\boldsymbol{W})$ & 0.0518 &\cellcolor{LightGreen} 0.0513 &\cellcolor{Pink2} 0.0520 &\cellcolor{LightGreen} 0.0625 & 0.0778 & 0.0651 &\cellcolor{Pink2} 0.0988\\
    NC3 &\cellcolor{LightGreen} 0.0174 & 0.0183 &\cellcolor{Pink2} 0.0254 &\cellcolor{LightGreen} 0.1419 & 0.1747 & 0.2115 &\cellcolor{Pink2} 0.2376\\
    NC4 &\cellcolor{LightGreen} 0.0000 &\cellcolor{LightGreen} 0.0000 &\cellcolor{LightPink2} 0.0001 &\cellcolor{LightGreen} 0.0000 & 0.0174 & 0.0691 &\cellcolor{DarkPink2} 0.1680\\
    \bottomrule
  \end{tabular}
\caption{Comparison of all NC metrics between trojaned and benign models for CIFAR-100 and GTSRB. $\boldsymbol{M}$ refers to the matrix of class means of the final-layer feature representations; $\boldsymbol{W}$ denotes the weight matrix of the final fully-connected layer. The percentage indicates the poison rate for the training dataset of the trojaned models. }
\label{table:NC Metrics for diff datasets}
\end{table*}

Figure~\ref{fig:NC2} tracks NC2, defined in Equation~\ref{eq: NC2}, which characterizes the convergence of both the final classifier weights and the training feature means to a simplex ETF. The red dashed line in the figures marks the point where the training error reaches zero and so the regime in which Neural Collapse is generally observed. For the equinorm property, shown in Figure~\ref{fig:NC2_equinorm_Means&Classifiers}, the benign model shows the most uniformity in norm across both feature means and classifier weights, and as the percentage of trojaned data increases, the equinorm property weakens. This trend is particularly profound in the classifier weights. We note that, for both class means and classifiers, the source of this deviation is actually that the norm of the target class is consistently \textit{smaller} than those of other classes. For the equiangular comparison, a similar, though smaller trend is observed. In Figure~\ref{fig:NC2_equiangular_Means&Classifiers}, the benign model reaches the smallest equiangular metric for both feature class means and classifiers. This further underscores the weakening of Neural Collapse due to trojaning.

NC3, defined in Equation~\ref{eq: NC3}, is shown in Figure~\ref{fig:NC3}.  Here, again, the metric converges to a smaller value for the benign model than either trojaned model and the disruption to Neural Collapse appears to increase with the proportion of training data poisoned with the trojan trigger.

But this pattern is not observed in every collapse metric for every dataset.  On CIFAR-10 we did not observe a significant difference between the benign and trojaned models for NC1, characterizing the variability across feature representations, or NC4, characterizing how close the network's decision rule is to a nearest neighbor rule in the final feature space, as they all go down to zero. However, collapse is noticeably weakened in these metrics in our experiments on CIFAR-100 and GTSRB, as shown in Figures~\ref{fig:NC1_GTSRB} and~\ref{fig:NC4_GTSRB} and Table~\ref{table:NC Metrics for diff datasets}.

Thus, while not every metric clearly differentiates trojan and benign models on every dataset, it is the case that benign models always show \textit{at least} the same degree of convergence as any trojaned model on every collapse metric. And, importantly, on every dataset we observe at least one metric where trojaned models show a significantly weaker degree of collapse than their benign counterpart--suggesting strong tension between Neural Collapse and an effective trojan trigger.

\section{Trojan Cleansing Method (ETF-FT)}
\subsection{Methodology} \label{ssec: Cleansing Method}

This tension motivates an algorithm for cleansing a network of any potential trojan triggers, which we name \textbf{ETF-FT}.  ETF-FT aims to take a fully trained, possibly trojaned neural network and remove trojan triggers, if any exist, while maintaining the network's good performance on clean test data.  The effects of a trojan attack are most apparent in~\cref{fig:NC2_equinorm_Means&Classifiers}, which shows that the weights of the final layer of a benign neural network exhibit a much greater degree of symmetry than those of an analogous trojaned network.  So, as specified in \cref{alg:Cleansing}, we proceed by first over-writing the weights of the final layer to a randomly generated simplex ETF, freezing those weights, and then fine-tuning the remaining model parameters on a small subset of clean data.  Notably, this process works well with even only limited data, and the fine-tuning data does not necessarily need to come from the original training set.

\begin{algorithm}[H]
\caption{Trojan Cleansing with Neural Collapse (ETF-FT)}
\label{alg:Cleansing}
\begin{algorithmic}
\STATE \textbf{Input:} Trojaned model $h_{\text{troj}}$, clean dataset $\mathbf{X}_{\text{clean}}$
\STATE \textbf{Step 1: Replace weight matrix with $\textbf{W}_{\text{ETF}}$}
\STATE \text{   }  Construct random ETF $\textbf{W}_{\text{ETF}}$ according to \cref{alg:etf_gen}
\STATE \text{   }  Replace the final layer weight matrix with $\textbf{W}_{\text{ETF}}$
\STATE \textbf{Step 2: Fine-tune the model}
\STATE \text{   }  Freeze the final layer weight matrix
\STATE \text{   }  Fine-tune on the clean dataset $\mathbf{X}_{\text{clean}}$
\STATE \textbf{Output:} Cleansed model $h_{\text{cleansed}}$
\end{algorithmic}
\end{algorithm}

\begin{algorithm}[H]
\caption{Constructing the $\mathbf{W}_{\text{ETF}}$ Matrix }
\label{alg:etf_gen}
\begin{algorithmic}
\STATE \textbf{Input:} Number of classes $K$, random orthogonal matrix $\mathbf{P} \in \mathbb{R}^{m \times K}$
\STATE \textbf{Initialize:} $\mathbf{I} \leftarrow \mathbf{I}_K$ (Identity matrix of size $K \times K$)
\STATE \textbf{Initialize:} $\mathbf{1} \leftarrow \mathbf{1}_{K \times K}$ (Matrix of ones of size $K \times K$)
\STATE Compute $\mathbf{W}_{\text{ETF}} \leftarrow \sqrt{\frac{K}{K - 1}} \times \mathbf{P} \times \left(\mathbf{I} - \frac{1}{K} \mathbf{1}\right)^T$
\STATE \textbf{Output:} Matrix $\mathbf{W}_{\text{ETF}}$
\end{algorithmic}
\end{algorithm}

ETF-FT is motivated by our results which show that trojans work, in part, by deforming the symmetric structure NC describes in the weights of the final layer.  By fixing those weights to be a random, but perfectly symmetric ETF our algorithm accomplishes two things:  first, it prevents the trojan deformation of the final layer and second, it resets the weights in a way known to be compatible with good classification.  This addresses a challenge present in many trojan cleansing techniques: the more you perturb the poisoned model the less effective the trigger will be, but the harder is it to recover accuracy on clean data.

\subsection{Experiment Details} \label{ssec: Comparison details}
We focus on two key metrics to evaluate the performance of each cleansing method: accuracy on clean data (ACC) and attack success rate (ASR). Accuracy reflects the model’s performance on clean data after cleansing, ideally showing no reduction compared to the original model. ASR, on the other hand, measures the proportion of poisoned samples classified as the attack target class. An effective cleansing method should reduce ASR to nearly zero.

We compare our method against several state-of-the-art cleansing algorithms that have demonstrated strong performance on the BackdoorBench~\cite{wu2022backdoorbench} leaderboard. As a baseline we run vanilla fine-tuning, with no trojan-aware modifications, on the same subset of clean data--these results are labeled FT in our results. 
We utilize the BackdoorBench repository \cite{wu2022backdoorbench} implementations of all attacks and cleansing methods besides our own. For experiments on PreAct-ResNet18, we adopt the default hyperparameters in the repository with a poisoning ratio 10\% for all of our experiments. 
We evaluate a range of backdoor attacks under the BackdoorBench setting, including BadNets-A2O~\cite{badnets}, TrojanNN~\cite{liu2018trojann}, ReFool~\cite{liu2020reflection}, WaNet~\cite{nguyen2021wanet}, BPP~\cite{wang2022bppattack}, CTRL~\cite{li2023embarrassingly}, Blended~\cite{chen2017targeted}, Blind~\cite{bagdasaryan2021blind}, Input-aware~\cite{nguyen2020input}, SIG~\cite{barni2019new}, and LIRA~\cite{doan2021lira}. Among these, Blind, LIRA, and SIG fail to achieve an ASR above 60\% while maintaining at least 70\% ACC, and are thus excluded from further analysis. Notably, none of the existing cleansing methods effectively mitigate ReFool or Blended attacks—ASR remains above 40\% when ACC is kept above 70\%. Therefore, we primarily focus on the remaining attacks, especially BadNets-A2O and TrojanNN. Experiments are conducted on two model architectures: PreAct-ResNet18~\cite{preactresnet} and ViT-B/16~\cite{alexey2020image}. 
We attempted to evaluate on ImageNet to test scalability but were unable to train trojaned models with both high ASR ($\geq50\%$) and reasonable ACC. Thus, these results are omitted.

\begin{table*}[b]
\scriptsize
\centering
\textbf{(a) PreAct-ResNet18}
\begin{tabular}{l c c | c c}
\toprule
\multirow{2}{*}{\textbf{Method}} &
\multicolumn{2}{c|}{\textbf{5\% Benign}} &
\multicolumn{2}{c}{\textbf{1\% Benign}} \\
 & ACC & ASR & ACC & ASR \\
\midrule
Original & 88.95 & 95.06  & 88.95 & 95.06 \\
FT & $91.63\pm0.18$&$10.03\pm7.23$& $90.20\pm0.22$&$22.04\pm25.38$ \\
NCleanse\cite{neuralCleanse} & $91.57\pm0.1$&$\textbf{0.98}\pm0.12$&$89.90\pm0.34$&$\textbf{1.12}\pm0.18$ \\
NAD\cite{NeuralAttentionDistillation} & $91.30\pm0.18$&$2.20\pm0.43$&$88.88\pm0.28$&$10.19\pm12.98$ \\
NPD\cite{zhu2024neural} & $85.71\pm3.59$ &$3.39\pm2.47$ &$83.26\pm3.59$&$5.26\pm3.82$ \\
FT-SAM\cite{zhu2023enhancing} & $\textbf{92.23}\pm0.12$&$2.37\pm0.81$&$\textbf{90.33}\pm0.15$&$8.12\pm11.86$ \\
I-BAU\cite{zeng2021adversarial} & $88.78\pm1.53$ & $5.53\pm6.81$ &$86.16\pm0.91$&$32.79\pm18.81$ \\
ETF-FT (ours) & $91.77\pm0.16$&$2.85\pm0.68$&$90.24\pm0.28$&$4.22\pm2.05$ \\
\bottomrule
\end{tabular}

\vspace{1.5em}

\textbf{(b) ViT-B/16}
\begin{tabular}{l c c | c c}
\toprule
\multirow{2}{*}{\textbf{Method}} &
\multicolumn{2}{c|}{\textbf{5\% Benign}} &
\multicolumn{2}{c}{\textbf{1\% Benign}} \\
 & ACC & ASR & ACC & ASR \\
\midrule
Original & 95.43 & 87.06 & 95.43 & 87.06 \\
FT & 96.06 & 20.46 & \textbf{95.74} & 36.87 \\
NCleanse\cite{neuralCleanse} & 96.08 & \textbf{0.28} & 95.61 & 2.70 \\
NAD\cite{NeuralAttentionDistillation} & 94.25 & 2.12 & 95.24 & 29.42 \\
NPD\cite{zhu2024neural} & 91.38 & 50.35 & 91.92 & \textcolor{red}{85.17} \\
FT-SAM\cite{zhu2023enhancing} & \textbf{96.38} & 1.93 & 94.29 & 6.16 \\
I-BAU\cite{zeng2021adversarial} & \textcolor{red}{25.96} & 10.87 & \textcolor{red}{12.91} & 68.68 \\
ETF-FT (ours) & 91.99 & 1.15 & 87.55 & \textbf{1.60} \\
\bottomrule
\end{tabular}
\caption{Comparison of cleansing methods against the \textbf{BadNets-A2O} attack on CIFAR-10 using 5\% and 1\% benign data. PreAct-ResNet18 results are averaged over 10 trials; ViT results are from a single trial.}
\label{tab:CleansingComparison_Badnet}
\end{table*}

We also evaluate the cleansing algorithm's performance on a vision transformer, due to the popularity and efficacy of transformer architectures.  However, many cleansing algorithms are not designed with transformers in mind.  We adapted algorithms to the new architecture where necessary, as in NPD, which inserts a new layer into the network, we chose to insert a linear layer before the final layer of the transformer encoder.  We also did light hyperparameter tuning over batch size, learning rate, scheduler and decay rate for each algorithm including our own.  Despite this, many algorithms show poor performance in this setting relative to their results on ResNet.  Many of these algorithms may be capable of better performance on transformers given careful adaptation, but in the trojan attack scenario the end user has outsourced network training to an adversary and so is unlikely to be a sophisticated deep learning engineer. As such, it is a significant strength of our algorithm that it can be applied to any classifier architecture without modification. 

\subsection{Cleansing Results} \label{ssec: cleansing results}

\textbf{BadNet Attack} \Cref{tab:CleansingComparison_Badnet} presents the cleansing results for the BadNet attack. When the cleansing algorithms are allowed to use 5\% of the training data to fine-tune, on the ResNet model, ETF-FT achieves competitive ASR while maintaining high accuracy on clean data relative to the other methods tested. In more restrictive conditions, where only 1\% of the training data is available, ETF-FT and Neural Cleanse maintain strong performance, while other methods show significant degradation. 

\textbf{Vision Transformers} Despite their state-of-the-art performance and growing popularity, vision transformers have rarely been examined in the context of backdoor cleansing performance.  This is in part due to the fact that their distinct architecture makes it difficult to adapt many common cleansing methods. \Cref{tab:CleansingComparison_Badnet} shows cleansing results for the BadNet attack applied to the ViT model, ETF-FT achieves very low ASR in both standard and limited data conditions, surpassing other methods except Neural Cleanse in the standard data case. This does come at the cost of a trade-off with in clean accuracy, which we believe can be mitigated, as discussed in \cref{ssec:future}.

\begin{table*}[b]
\scriptsize
\centering

\textbf{(a) PreAct-ResNet18}

\begin{tabular}{l c c | c c}
\toprule
\multirow{2}{*}{\textbf{Method}} &
\multicolumn{2}{c|}{\textbf{5\% Benign}} &
\multicolumn{2}{c}{\textbf{1\% Benign}} \\
 & ACC & ASR & ACC & ASR \\
\midrule
Original & 91.17 & 99.90 & 91.17 & 99.90 \\
FT & $91.92\pm0.15$&$3.12\pm3.14$& $90.60\pm0.37$ & $3.75\pm6.41$ \\
NCleanse\cite{neuralCleanse} & $91.49\pm0.45$ & $41.07\pm48.04$ & $90.50\pm1.05$ & $70.67\pm44.64$ \\
NAD\cite{NeuralAttentionDistillation} & $91.36\pm0.28$ & $1.51\pm0.37$ & $88.39\pm0.89$ & $2.99\pm1.94$ \\
NPD\cite{zhu2024neural} & $82.65\pm2.14$ & $16.34\pm11.34$ & $86.87\pm0.13$ & $43.03\pm1.68$ \\
FT-SAM\cite{zhu2023enhancing} & $90.78\pm0.21$ & $8.10\pm4.62$ & $87.52\pm0.60$ & $12.23\pm19.80$ \\
I-BAU\cite{zeng2021adversarial} & $91.83\pm0.12$ & $\textcolor{red}{82.22}\pm1.73$ & $\textbf{91.73}\pm0.25$ & $\textcolor{red}{99.91}\pm0.01$ \\
ETF-FT (ours) & $\textbf{92.39}\pm0.24$ & $\textbf{0.04}\pm0.05$ & $91.01\pm0.47$ & $\textbf{0.00}\pm0.00$ \\
\bottomrule
\end{tabular}

\vspace{1.5em}

\textbf{(b) ViT-B/16}

\begin{tabular}{l c c | c c}
\toprule
\multirow{2}{*}{\textbf{Method}} &
\multicolumn{2}{c|}{\textbf{5\% Benign}} &
\multicolumn{2}{c}{\textbf{1\% Benign}} \\
 & ACC & ASR & ACC & ASR \\
\midrule
Original & 96.67 & 100.00 & 96.67 & 100.00 \\
FT & 95.94 & \textcolor{red}{100.00} & 95.98 & \textcolor{red}{100.00} \\
NCleanse\cite{neuralCleanse} & 96.80 & \textbf{2.33} & 96.52 & 40.96 \\
NAD\cite{NeuralAttentionDistillation} & 89.99 & \textcolor{red}{100.00} & 96.40 & \textcolor{red}{100.00} \\
NPD\cite{zhu2024neural} & 96.69 & \textcolor{red}{100.00} & 96.50 & \textcolor{red}{100.00} \\
FT-SAM\cite{zhu2023enhancing} & \textbf{96.98} & \textcolor{red}{100.00} & \textbf{96.82} & \textcolor{red}{100.00} \\
I-BAU\cite{zeng2021adversarial} & \textcolor{red}{36.10} & 9.74 & \textcolor{red}{36.70} & \textbf{0.41} \\
ETF-FT (ours) & 95.34 & 10.76 & 95.04 & 10.11 \\
\bottomrule
\end{tabular}

\caption{Comparison of cleansing methods on \textbf{TrojanNN}-attacked CIFAR-10 models using 5\% and 1\% benign data. PreAct-ResNet18 results are averaged over 10 trials; ViT results are from a single trial.}
\label{tab:CleansingComparison_TrojanNN}

\end{table*}

\textbf{TrojanNN Attack} Alongside the standard BadNet attack, we consider the more sophisticated TrojanNN attack~\cite{liu2018trojann} which identifies specific vulnerable neurons in a pre-trained network and reverse engineers a subtle trigger to target them. As shown in \cref{tab:CleansingComparison_TrojanNN}, ETF-FT is particularly well suited to cleansing this style of trojan.  On ResNets, ETF-FT completely eliminates the trigger, achieving an ASR of nearly 0 across 10 trials, while maintaining competitive accuracy with the other cleansing methods. On ViT, many methods, implemented out of the box, fail completely to eliminate the trigger, while ETF-FT achieves an ASR of 10\% with only a 1.5 point drop in clean accuracy. I-BAU algorithm is able to almost entirely eliminate the trigger, but in doing so destroyed the model's accuracy on clean data.

\textbf{Other Attacks} We also evaluate additional attacks, as detailed in \cref{ssec: Comparison details}. Results for WaNet, BPP, and Input-aware are presented in \cref{tab: CleansingComparison_Other}. ETF-FT demonstrates competitive performance against WaNet and BPP, and achieves state-of-the-art results on the Input-aware attack.

\begin{table*}[h]
    \centering
    \scriptsize
    \begin{tabular}{lcc|cc|cc}
    \toprule
    \multirow{2}{*}{\textbf{Method}} &
    \multicolumn{2}{c|}{\textbf{WaNet}} &
    \multicolumn{2}{c|}{\textbf{BPP}} &
    \multicolumn{2}{c}{\textbf{Input-aware}} \\
    & \textbf{ACC} & \textbf{ASR} & \textbf{ACC} & \textbf{ASR} & \textbf{ACC} & \textbf{ASR} \\
    \midrule
    Original    & 93.69 & 97.12 & 94.58 & 99.98 & 94.25 & 99.36 \\
    FT          & 92.46 & 0.79 & 92.45 & 1.86 & 92.36 & 1.21 \\
    NCleanse\cite{neuralCleanse} 
                & 92.85 & 39.19 & 87.51 & 21.57 & 86.07 & 12.79 \\
    NAD\cite{NeuralAttentionDistillation} 
                & 91.68 & 0.88 & \textcolor{red}{31.55} & 8.95 & \textcolor{red}{32.63} & 5.85 \\
    NPD\cite{zhu2024neural} 
                & 87.68 & 2.44 & 85.62 & \textcolor{red}{84.38} & 91.85 & 1.29 \\
    FT-SAM\cite{zhu2023enhancing} 
                & \textbf{93.60} & \textbf{0.69} & 89.96 & \textbf{1.74} & 91.14 & 3.57 \\
    I-BAU\cite{zeng2021adversarial} 
                & 91.85 & 4.24 & 91.84 & 5.21 & 92.08 & 10.35 \\
    ETF-FT (ours) 
                & 93.29 & 0.88 & \textbf{93.49} & 16.60 & \textbf{93.38} & \textbf{0.96} \\
    \bottomrule
    \end{tabular}
    \caption{Comparison of cleansing methods on \textbf{WaNet, BPP, Input-aware}-attacked CIFAR-10 models using 5\% benign data. Results are on PreAct-ResNet18, averaged over 10 trials. Standard deviation in \cref{tab:other_attacks_std} in appendix \ref{appendix:std}.}
    \label{tab: CleansingComparison_Other}
\end{table*}

\subsubsection{Robustness}

\begin{table*}[h]
  \centering
  \begin{tabular}{l c c }
    \toprule
    \textbf{Method} & \textbf{ACC} & \textbf{ASR} \\
    \midrule
    Original & 88.95 & 95.06 \\
    FT & $81.66\pm0.39$ &$1.76\pm0.48$\\
    NCleanse\cite{neuralCleanse} & $81.80\pm0.38$&$1.72\pm0.38$\\
    NAD\cite{NeuralAttentionDistillation} & $67.14\pm1.96$&$3.10\pm1.68$\\
    NPD\cite{zhu2024neural} & $83.10\pm3.64$&$5.97\pm4.59$\\
    FT-SAM\cite{zhu2023enhancing} & $85.23\pm0.77$&$\textbf{1.21}\pm0.48$\\
    I-BAU\cite{zeng2021adversarial} &$86.65\pm1.38$&	$28.26\pm17.91$ \\
    ETF-FT(ours) & $\textbf{87.99}\pm0.49 $&$3.37\pm2.80$\\ 
    \bottomrule
  \end{tabular}
  \caption{\textbf{Data Corruption}: Comparison of cleansing algorithms with \textbf{1\%} corrupted (under random erasure) data on PreAct-ResNet18 trained on CIFAR-10 with a BadNet attack,  averaged over 5 trials. }
  \label{tab:RandomErasing}
\end{table*}

The above results show that ETF-FT performs well with access to only a small amount of training data, but it is likely that the user of a trojaned model may not have access to any data from the original training distribution at all.  We simulated distribution shift in two ways: by imbalancing the class distribution and by applying random erasures within image patches. Results for these datasets, applied to the ResNet BadNet poisoned models, are shown in \cref{tab:RandomErasing} and \cref{tab:Imbalance} (in appendix \ref{appendix:results}).  In both settings, ETF-FT excels at maintaining the model's clean accuracy along with competitive ASR relative to other methods.  Neural Cleanse, which had strong results with access to the original training data, performs significantly better than all other methods on the imbalanced data, but its clean accuracy suffers on the corrupted data.  This is due to the fact that its pruning method relies on identifying important neurons based on their activations on the available data--so it is vulnerable to accidentally pruning neurons associated with valid features which aren't represented in the fine-tuning dataset.

\subsubsection{More Sophisticated Trigger}
\begin{figure}[h]
    \centering
    \includegraphics[width=0.25\linewidth]{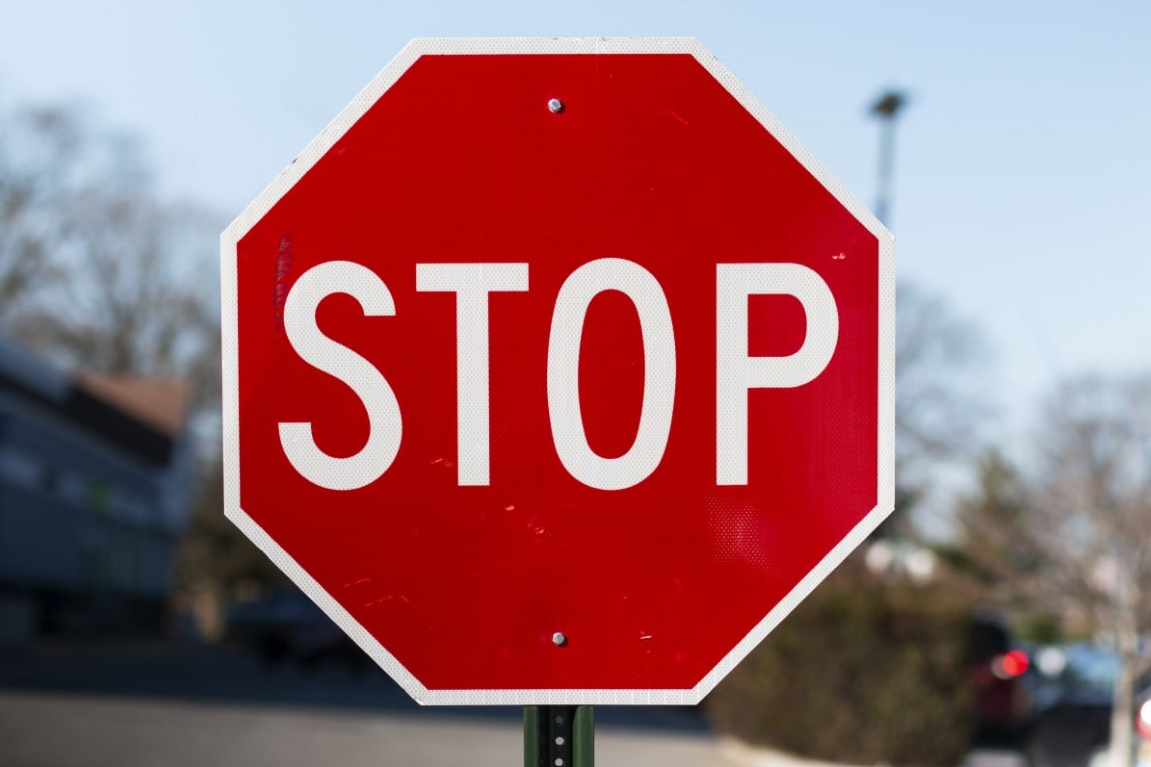}\hspace{1pt}%
    \includegraphics[width=0.25\linewidth]{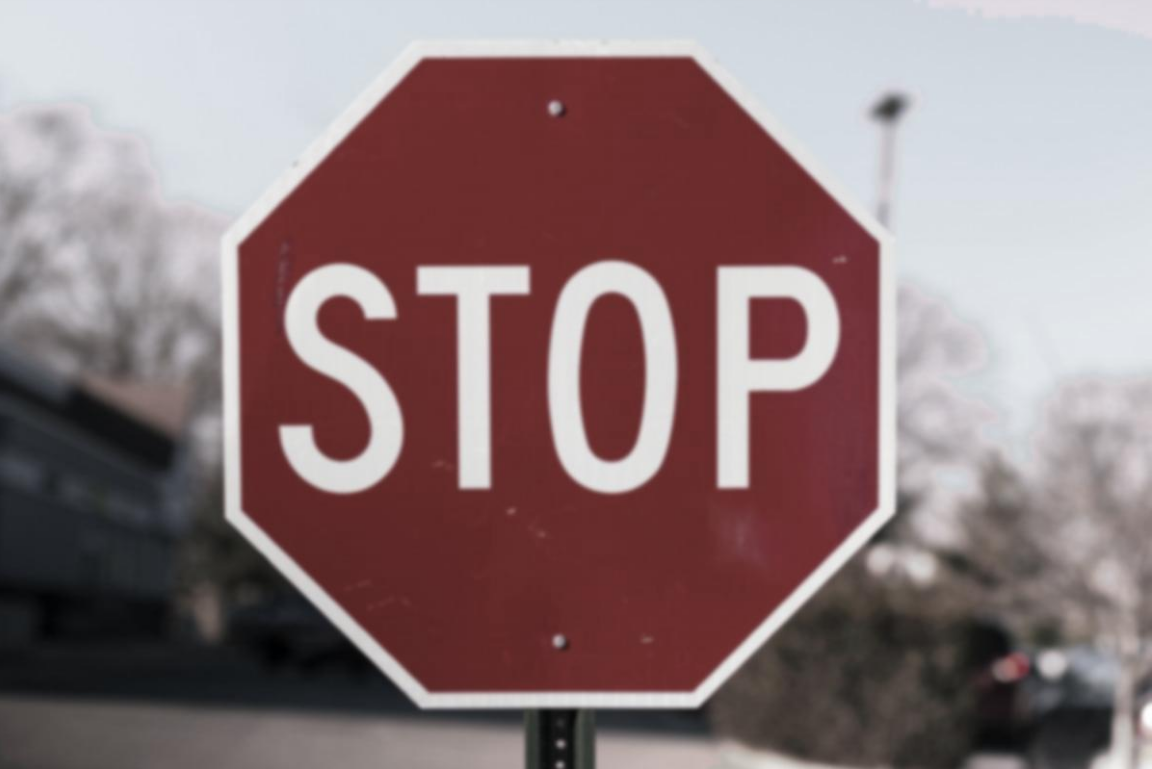}\hspace{1pt}%
    \includegraphics[width=0.25\linewidth]{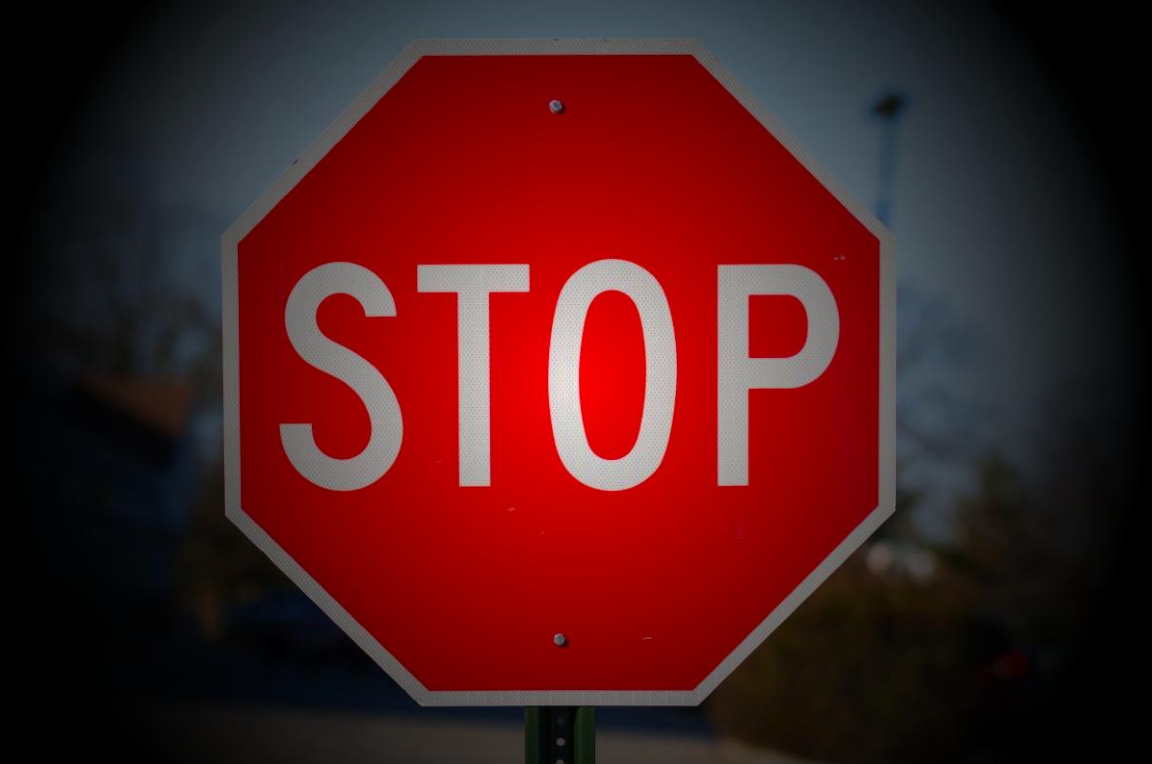}

    \captionof{figure}{Instagram filter effects (Original, Gotham, Lomo)}
    \label{fig:Instagram Filter}
\end{figure}

To thoroughly evaluate the performance of ETF-FT against Trojan attacks with more sophisticated triggers beyond the basic yet powerful patch trigger, we conduct experiments using Instagram filter triggers \cite{trojai_instagram_xforms}. A demonstration of the filter effects is presented in \cref{fig:Instagram Filter}, where the Gotham filter introduces blurring and a gray tone, while the Lomo filter creates a vignette effect. As shown in \cref{tab:CleansingComparison_Instagram}, ETF-FT remains effective even against this more subtle type of trigger.

\begin{table*}[h]
    \centering
    \begin{tabular}{lcc| c c}
        \toprule
        \multirow{2}{*}{\textbf{Method}} &
        \multicolumn{2}{c|}{\textbf{Gotham}} &
        \multicolumn{2}{c}{\textbf{Lomo}} \\
        & \textbf{ACC} & \textbf{ASR} & \textbf{ACC} & \textbf{ASR} \\
        \midrule
        Original & 93.66 & 100  & 93.21 & 100 \\
        FT & $89.84 \pm 0.35$ & $24.5\pm21.58$ & $\textbf{89.92}\pm0.22$ & $3.6\pm2.11$ \\
        NCleanse\cite{neuralCleanse} &$84.54\pm6.05$ & $\textcolor{red}{70.49}\pm28.77$ & $83.1\pm5.18$ & $22.68\pm38.73$\\
        NAD\cite{NeuralAttentionDistillation} &$\textcolor{red}{29.48}\pm2.61$ & $4.14\pm4.03$ &$\textcolor{red}{29.52}\pm2.48$ & $\textbf{0.0}\pm0.0$ \\
        NPD\cite{zhu2024neural} & $87.21\pm2.53$ & $\textbf{3.87}\pm5.29$ & $87.68\pm3.62$ & $20.82\pm18.39$ \\
        FT-SAM\cite{zhu2023enhancing} & $85.94\pm1.42$ & $10.59\pm12.85$ & $86.93\pm1.22$ & $1.84\pm1.03$ \\
        I-BAU\cite{zeng2021adversarial} & $\textbf{90.34}\pm0.86$ & $39.12\pm27.04$ & $89.86\pm1.56$ & $8.84\pm5.58$ \\
        ETF-FT(ours) &$85.4\pm0.54$ & $14.9\pm20.7$&$85.33\pm0.52$&$2.28\pm1.29$\\
        \bottomrule
    \end{tabular}
    \caption{Performance of cleansing methods on BadNets-trojaned CIFAR-10 models using \textbf{Gotham and Lomo Instagram filter} triggers. Cleansing is performed with 5\% clean data. Results are reported on PreAct-ResNet18, averaged over 10 trials.}
    \label{tab:CleansingComparison_Instagram}
\end{table*}

\subsubsection{Adaptive Attack}

Consider a scenario where the adversary is aware that trojan attacks undermine the phenomenon of Neural Collapse. Thus, they train the model with the final layer's weights fixed as an ETF, effectively replicating the conditions under which Neural Collapse occurs perfectly. ETF-FT demonstrates robust performance even in the face of adaptive attacks. As evidenced in \cref{tab:AdaptiveAttack}, the ASR is only 90.65, which is lower than the ASR observed in other similar settings. This supports our observation that trojan attacks are incompatible with the Neural Collapse mechanism. Notably, ETF-FT not only removes the trojans but also enhances the model's Accuracy.

\begin{table*}[h]
\centering
  \begin{tabular}{c c c }
    \toprule
      & ACC & ASR \\
    \midrule
    Original ($W$ fixed as ETF) & 87.68 & 90.65  \\
    ETF-FT (5\% clean data) & $91.48 \pm0.18$&$2.56\pm0.67$\\
    ETF-FT (1\% clean data) & $89.91\pm0.3$&$4.61\pm4.22$\\
    \bottomrule
  \end{tabular}
  \captionof{table}{\textbf{Adaptive Attack}: Performance of ETF-FT on models trained with \textbf{final layer fixed as ETF}. Results are on BadNets-trojaned CIFAR-10 using PreAct-ResNet18, averaged over 10 trials.  }
  \label{tab:AdaptiveAttack}
\end{table*} 

\subsubsection{Cleansing Without Overtraining}
Neural Collapse occurs most strongly in models trained well past zero training error. In line with this, our other experiments used models trained for 350 epochs, despite reaching zero training error at approximately 120 epochs. However, in practical scenarios, models may not always be overtrained. To examine ETF-FT's reliance on model overtraining we also test it on models trained only until the training error reaches zero. As shown in \cref{tab:NotOvertrained}, ETF-FT maintains strong performance in this case, achieving comparable results to its performance on overtrained models.

\begin{table*}[h]
  \centering
  \begin{tabular}{c c c }
    \toprule
      & ACC & ASR \\
    \midrule
    Original & 90.66 & 94.78  \\
    ETF-FT (5\% clean data) & $91.3\pm0.1$&$0.86\pm0.23$\\
    ETF-FT (1\% clean data) & $89.54\pm0.31$&$0.72\pm0.23$\\
    \bottomrule
  \end{tabular}
  \captionof{table}{\textbf{No Overtraining}: Performance of ETF-FT on trojaned model only trained for \textbf{120 epochs}. Results are on BadNets-trojaned CIFAR-10 using PreAct-ResNet18, averaged over 10 trials.  }
  \label{tab:NotOvertrained}
\end{table*}

\subsubsection{Complex Datasets}
We further assess the robustness of ETF-FT on two challenging benchmarks—CIFAR-100 (100 classes) and GTSRB (43 traffic-sign classes with non-uniform image sizes)—to probe generalization under greater class granularity and real-world variability. As summarized in \Cref{tab:cifar100_results}, ETF-FT consistently maintains a high clean accuracy while substantially reducing ASR in both datasets, maintaining a high competency among the cleansing methods.

\begin{table*}[h]
\scriptsize
\centering
\renewcommand{\arraystretch}{1.1}
\setlength{\tabcolsep}{5pt}
\begin{tabular}{lcccc}
\toprule
\multirow{2}{*}{\textbf{Method}} & 
\multicolumn{2}{c}{\textbf{CIFAR-100}} &
\multicolumn{2}{c}{\textbf{GTSRB}} \\
\cmidrule(lr){2-3} \cmidrule(lr){4-5}
& \textbf{ACC} & \textbf{ASR} & \textbf{ACC} & \textbf{ASR} \\
\midrule
Original & 70.69 & 87.25 & 95.96 & 89.17\\

FT & \textbf{70.80} $\pm$ 0.14 & 47.02 $\pm$ 19.59 & 97.39 $\pm$ 0.16 & 0.01 $\pm$ 0.02 \\
NCleanse \cite{neuralCleanse} & 69.96 $\pm$ 0.54 & \textbf{0.05} $\pm$ 0.03 & 97.05 $\pm$ 0.27 & 0.00 $\pm$ 0.01 \\

NAD\cite{NeuralAttentionDistillation} & 70.61 $\pm$ 0.21 & 42.18 $\pm$ 25.56 & 97.10 $\pm$ 0.27 & 0.00 $\pm$ 0.01 \\

NPD\cite{zhu2024neural} & 60.39 $\pm$ 0.72 & 0.21 $\pm$ 0.11 & 95.29 $\pm$ 0.30 & 0.58 $\pm$ 0.34 \\
FT-SAM\cite{zhu2023enhancing} & 68.51 $\pm$ 0.30 & 16.33 $\pm$ 15.13 & \textbf{97.98} $\pm$ 0.15 & 0.01 $\pm$ 0.01 \\

I-BAU\cite{zeng2021adversarial} & 68.63 $\pm$ 0.74 & 3.34 $\pm$ 3.81 & 95.06 $\pm$ 0.67 & 0.05 $\pm$ 0.06 \\


ETF-FT (ours) & 66.09 $\pm$ 0.39 & 0.36 $\pm$ 0.13 & 97.90 $\pm$ 0.19 & \textbf{0.00} $\pm$ 0.00 \\
\bottomrule
\end{tabular}
\caption{\textbf{CIFAR100 \& GTSRB}: Comparison of cleansing methods on on BadNet attacked PreAct-ResNet18 models with 5\% clean data. Averaged over 10 trials.}
\label{tab:cifar100_results}
\end{table*}

\section{Conclusion}
In this paper, we explore how the asymmetry induced by a trojan trigger in a neural network disrupts the phenomenon of Neural Collapse. We provide experimental evidence showing that trojan attacks significantly weaken NC based on standard metrics. Leveraging these insights, we propose a novel method for trojan cleansing: by fixing the network’s final layer to a simplex ETF—an ideal structure predicted by NC—and fine-tuning on a small amount of clean data. Our results show that this approach effectively removes the trojan trigger while maintaining the network’s accuracy on clean test data without requiring knowledge of the trigger type, the target class, or access to the original training data. Moreover, our method can be applied to any standard architecture with no modification.

\subsection{Future Work}
\label{ssec:future}
This work suggests a variety of interesting future directions.  
Given the significant experimental evidence of the connection between trojans and Neural Collapse, developing a theoretical framework as well as building a trojan detection method ae important areas of future work. 

 In this work, we used a randomly generated ETF as the starting point in our cleansing strategy; characterizing the impact of the  initial ETF selection as well as optimizing this are other interesting areas of investigation. 

 This paper focuses on image classification, but trojans have also been proven effective in other domains such as language and  audio.  Extending our analysis and cleansing mechanism to these domains is an interesting direction for future work.   

\bigskip





\newpage

\begin{appendices}

\section{More Cleansing Results}
\label{appendix:results}

\begin{table*}[h]
  \centering
  \begin{tabular}{c c c}
    \toprule
    Method & ACC & ASR  \\
    \midrule
    Original & 88.95 & 95.06   \\
    FT & $86.01\pm1.59$&$22.95\pm19.81$\\
    NCleanse\cite{neuralCleanse} & $\textbf{88.43}\pm0.29$&$\textbf{1.19}\pm0.27$\\
    NAD\cite{NeuralAttentionDistillation} & $83.41\pm1.87$&$3.17\pm0.72$\\
    NPD\cite{zhu2024neural} & $83.10\pm3.64$&$5.97\pm4.59$\\
    FT-SAM\cite{zhu2023enhancing} & $87.20\pm0.54$&$7.17\pm3.69$\\
    I-BAU\cite{zeng2021adversarial} &$85.97\pm1.43$&	$49.55\pm12.15$ \\
    ETF-FT(ours) & $88.16\pm0.56$&$7.18\pm2.06$\\ 
    \bottomrule
  \end{tabular}
  \caption{\textbf{Data Imbalance}: Comparison of cleansing algorithms with \textbf{1\%} imbalanced clean training data on BadNet trojaned ResNet models(\%). Results are averaged over 5 random trials with mean$\pm$std. }
  \label{tab:Imbalance}
\end{table*}

\section{Standard Deviation of \Cref{tab: CleansingComparison_Other}} \label{appendix:std}

We append the standard deviation of \cref{tab: CleansingComparison_Other} here for reference.

\begin{table*}[h]
\centering
\begin{tabular}{l|cc|cc|cc}
\toprule
\multirow{2}{*}{Method} 
& \multicolumn{2}{c|}{WaNet} 
& \multicolumn{2}{c|}{BPP} 
& \multicolumn{2}{c}{Input-aware} \\
& ACC & ASR & ACC & ASR & ACC & ASR \\
\midrule
Original      
& --        & --        & --        & --        & --        & --        \\
FT            
& 0.09  & 0.21  & 0.23  & 0.35  & 0.28  & 0.28  \\
NCleanse\cite{neuralCleanse} 
& 0.76  & 47.30 & 3.67  & 39.21 & 2.82  & 28.86 \\
NAD\cite{NeuralAttentionDistillation} 
& 0.31  & 0.23  & 2.57  & 5.04  & 4.39  & 3.74  \\
NPD\cite{zhu2024neural} 
& 4.22  & 2.40  & 5.74  & 11.56 & 1.88  & 2.02  \\
FT-SAM\cite{zhu2023enhancing} 
& 0.13  & 0.11  & 0.65  & 0.50  & 0.49  & 1.48  \\
I-BAU\cite{zeng2021adversarial}
& 0.38  & 3.51  & 0.73  & 31.21 & 0.57  & 11.70 \\
ETF-FT (ours)
& 0.13  & 0.19  & 0.15  & 9.54  & 0.18  & 0.26  \\
\bottomrule
\end{tabular}
\caption{
\textbf{Standard deviation} for the comparison of cleansing methods on WaNet, BPP, and Input-aware attacked CIFAR-10 models.  
Results are calculated over 10 trials.
}
\label{tab:other_attacks_std}
\end{table*}

\section{Code and Instructions}
\label{appendix: code and instructions}

Our implementation is integrated into the BackdoorBench~\cite{wu2022backdoorbench} repository. The core implementation of our method is provided in the file \texttt{etf-ft.py}, which should be placed under \texttt{BackdoorBench/defense/}. The directory \texttt{etf-ft/} contains configuration files and hyperparameter settings used for our experiments and should be placed under \texttt{BackdoorBench/config/defense/}. Note that ETF-FT relies on the \texttt{AdamW} optimizer and \texttt{ExponentialLR} scheduler, which must be added to the BackdoorBench environment if not already available. Once configured, ETF-FT can be executed using the same command interface as other defense methods in the framework.

For experiments involving the Instagram filter trigger, the transformation is implemented in \texttt{filter.py}, based on the module proposed in~\cite{trojai_instagram_xforms}, and should be placed in \texttt{BackdoorBench/utils/bd\_img\_transform/}.

\section{Experimental Details}
\label{appendix: experimental details}

ETF-FT employs the \texttt{AdamW} optimizer in conjunction with an \texttt{ExponentialLR} learning rate scheduler. The specific learning rate and decay factor (\texttt{gamma}) are tuned based on the model architecture.

For \textbf{PreAct-ResNet18} (on CIFAR-10/CIFAR-100/GTSRB), we use:
\begin{itemize}
\item Learning rate: \texttt{0.0001}
\item Learning rate scheduler: \texttt{ExponentialLR}
\item \texttt{gamma}: \texttt{0.95}
\item Optimizer: \texttt{AdamW}
\end{itemize}

For \textbf{ViT on CIFAR-10}, we use:
\begin{itemize}
\item Learning rate: \texttt{0.001}
\item Learning rate scheduler: \texttt{ExponentialLR}
\item \texttt{gamma}: \texttt{0.85}
\item Optimizer: \texttt{AdamW}
\end{itemize}




\end{appendices}

\newpage

\bibliography{sn-bibliography}

\end{document}